\pdfoutput=1

\documentclass[11pt]{article}

\usepackage[]{emnlp2021}

\usepackage{times}
\usepackage{latexsym}

\usepackage[T1]{fontenc}

\usepackage[utf8]{inputenc}

\usepackage{microtype}
\usepackage{tabularx}
\usepackage{footnote}
\usepackage{scalerel,xparse}
\NewDocumentCommand\emojiLaugh{}{
        \includegraphics[scale=0.12]{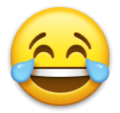}
}
\usepackage{placeins}
\usepackage{multirow}
\usepackage{float}
\usepackage[normalem]{ulem}
\useunder{\uline}{\ul}{}
\usepackage{caption}

\title{User Guide for KOTE: Korean Online Comments Emotions Dataset}

\author{Duyoung Jeon \and Junho Lee \and Cheongtag Kim \\
  Department of Psychology, Seoul National University\\
  Seoul, Republic of Korea\\
  \texttt{wuju1201@gmail.com}, \texttt{\{smbslt3, ctkim\}@snu.ac.kr} \\}

\begin{document}
\maketitle
\begin{abstract}
Sentiment analysis that classifies data into positive or negative has been dominantly used to recognize emotional aspects of texts, despite the deficit of thorough examination of emotional meanings. Recently, corpora labeled with more than just valence are built to exceed this limit. However, most Korean emotion corpora are small in the number of instances and cover a limited range of emotions. We introduce KOTE dataset\footnote{\url{https://github.com/searle-j/KOTE}}. KOTE contains 50k (250k cases) Korean online comments, each of which is manually labeled for 43 emotion labels or one special label (NO EMOTION) by crowdsourcing (Ps = 3,048). The emotion taxonomy of the 43 emotions is systematically established by cluster analysis of Korean emotion concepts expressed on word embedding space. After explaining how KOTE is developed, we also discuss the results of finetuning and analysis for social discrimination in the corpus.
\end{abstract}

\section{Introduction}

Sentiment analysis that classifies texts into positive or negative has been the most widely used method to analyze the emotional aspect of texts. Although sentiment analysis is simple, feasible, and useful in various situations, the need for more sophisticated emotions beyond just valence for text analysis is emerging. This is due to the advent of powerful language models that can accommodate complicatedly labeled data and the recent advancement in computing power.

\begin{table}

    \begin{tabularx}{\linewidth}{X}\hline
      \centering
      \textbf{Text}
    \end{tabularx}
    
    \begin{tabularx}{\linewidth}{X}\hline
      you silly cat made a fuss just because you didn’t want to take a bath?? LOL \emojiLaugh\\ 
    \end{tabularx}
    
    \begin{tabularx}{\linewidth}{X}\hline
      \centering
      \textbf{Labels}
    \end{tabularx}
    
    \begin{tabularx}{\linewidth}{X}\hline
      {\small \textbf{rater 1} \textit{preposterous, attracted, care, happiness}}\\
      {\small \textbf{rater 2} \textit{preposterous, attracted, embarrassment, realization}}\\
      {\small \textbf{rater 3} \textit{preposterous, interest, embarrassment, irritation, dissatisfaction}}\\
      {\small \textbf{rater 4} \textit{preposterous}}\\
      {\small \textbf{rater 5} \textit{attracted, interest, excitement}}\\\hline
    \end{tabularx}

\caption{\raggedright A raw example in KOTE.}
\label{tab:1}

\end{table}

The demand for an emotion analysis tool for the Korean language is high. However, most Korean emotion corpora are small in the number of instances and have coarse emotion taxonomies that cover only a limited range of emotions. As a result, GoEmotions \citep{demszky-etal-2020-goemotions}, an English dataset that is large (58k instances) and has a fine-grained emotion taxonomy (27 emotions or neutral), is widely used for emotion analysis for Korean text using machine translation, despite the imperfect translation quality. The Korean language model trained with translated GoEmotions is downloaded up to hundreds of thousands of times a month in Hugging Face\footnote{\url{https://huggingface.co/monologg/bert-base-cased-goemotions-original}}.

However, emotions are strongly related to culture since they are products of culture-specific schema. Accordingly, emotion taxonomies representing underlying emotion structures vary across cultures \cite{mesquita1992cultural} and the variation even holds for basic emotions \cite{gendron2014perceptions}. This demonstrates the need to create a culturally relevant dataset that is labeled with a culturally relevant emotion taxonomy.

To create a culturally relevant database, we developed KOTE (Korean Online That-gul\footnote{‘That-gul’ or ‘Daet-gul’ is a Korean word that refers to ‘online comment’.} Emotions), a large language dataset of 50k Korean online comments labeled for 43 emotions. The online comments in KOTE are collected from 12 different platforms of various domains (\textit{news, online community, social media, e-commerce, video platform, movie review, microblog, and forum}). The 43 emotions befitting to the Korean language are derived from the clustering results of Korean words that refer to emotion concepts. \textbf{Table~\ref{tab:1}} shows a raw example in KOTE.

The purpose of this study is twofold. The first purpose suggests a new emotion taxonomy that is suitable to the Korean language in general. The second purpose builds KOTE with the new taxonomy. We also finetuned the pretrained \texttt{KcELECTRA} (Korean comment ELECTRA; \citealp{clark2020electra, lee2021kcelectra}) model with KOTE. This achieves a better performance than the existing model trained with translated GoEmotions (F1-scores are 0.56 versus 0.41). There is much room to improve since the results are not tuned. A diversity of strategies can possibly be applied on the raw data according to the individual purpose of an analyst because KOTE is fully open and contains rich information.

\section{Related Work}
\begin{table*}[hbt!]
\centering
\begin{tabular}{c|ccc}
\hline
\textbf{Dataset}                                                                                                     & \textbf{Unit} & \textbf{\# of instances} & \textbf{Label dimension} \\ \hline
\begin{tabular}[c]{@{}c@{}}Korean Emotion\\ Words Inventory\\ {\small \citep{park2005making}}\end{tabular}                       & word          & 434                      & 4                        \\ \hline
\begin{tabular}[c]{@{}c@{}}Korean Emotion\\ Vocabulary Taxonomy\\ {\small \citep{sohn2012korean}}\end{tabular}                   & word          & 504                      & 11                       \\ \hline
\begin{tabular}[c]{@{}c@{}}KOSAC\\ {\small \citep{jang-etal-2013-kosac}}\end{tabular}                                                  & sentence      & 7.7k                     & 2*                       \\ \hline
\begin{tabular}[c]{@{}c@{}}NSMC\\ {\small \citep{nsmc2015}}\end{tabular}                                                         & sentence      & 200k                     & 1                        \\ \hline
\begin{tabular}[c]{@{}c@{}}KNU SentiLex\\ {\small \citep{park2018knu}}\end{tabular}                                           & n-gram        & 14k                      & 1                        \\ \hline
\begin{tabular}[c]{@{}c@{}}Korean Continuous Dialogue Dataset\\ with Emotion Information\\ {\small \citep{keti2020continuous}}\end{tabular} & dialogue      & 10k                      & 7                        \\ \hline
\begin{tabular}[c]{@{}c@{}}Korean One-off Dialogue Dataset\\ with Emotion Information \\ {\small \citep{keti2020oneoff}}\end{tabular}   & sentence      & 38k                      & 7                        \\ \hline
\begin{tabular}[c]{@{}c@{}}Emotional Dialogue Corpus \\ {\small \citep{aihub2021dialogue}}\end{tabular}                                  & dialogue      & 15k                      & 60                       \\ \hline
\end{tabular}

\caption{Korean emotion text datasets.\\ {\small * KOSAC contains far more plentiful information, but two dimensions are closely related to emotion (\textit{polarity} and \textit{intensity}).}}
\label{tab:2}
\end{table*}

\subsection{Emotion Taxonomy}

Constructing an emotion corpus requires an appropriate emotion taxonomy by which the texts are labeled. To find the appropriate emotion taxonomy, constructing an emotion words dataset must precede to obtain all available emotions each of which is treated as a candidate to be included in the taxonomy.

Thus, the very first question is how to identify the types of emotion. Vocabulary representing emotions can be used to this end. In traditional approaches, the distinction between emotion and nonemotion is determined by human rating. \citet{shields1984distinguishing} attempted to conceptualize \textit{emotionality} by asking participants to categorize 60 feeling words (\textit{happy, curious, hungry, etc.}) into emotion or nonemotion words. \citet{clore1987psychological} measured the emotionality of 585 feeling words by asking participants to rate their confidence in a 4-point scale of how emotional each word is. Apart from the survey approaches, the emotionality can be determined by experts. \citet{averill1975semantic} recruited graduate students to scrutinize approximately 18k psychological concepts and concluded that 717 words contained emotionality. For an example of a Korean study, \citet{sohn2012korean} collected 65k Korean words from a variety of text sources and manually checked their properties to confirm 504 emotional expressions.

The next question after identifying the emotion words is how to transform the words into a mathematically analyzable form. One popular way is vectorization, which imposes vector-valued information on words by a certain measure. One classic way of the vectorization is by human rating, which is conducted by asking human annotators to rate each word in a few scales designed by researchers. For example, \citet{block1957studies} asked the participants to rate fifteen emotion words in twenty 7-point scales (e.g., \textit{good-bad, active-passive, tense-relaxed}). Similarly, \citet{sohn2012korean} vectorized 504 emotion words in eleven 10-point emotion scales (e.g., \textit{joy, anger, sadness}). \citet{park2005making} rated emotion words in four scales (i.e., \textit{prototypicality, familiarity, valence, and arousal}).

The vector of a word can be indirectly estimated via rating similarity (or distance) among words. \citet{storm1987taxonomic} utilized a sorting method to extract co-occurrence information from emotion words. \citet[p.75]{cowen2019mapping} suggested that a pseudorandom assignment for similarity rating is sufficient to embed the local similarity of 600 emotion words.

The last question is how to uncover an adequate structure of the emotion words using the information. ‘How many emotions are there?’ has always been one of the biggest and the most mesmerizing questions in the field of emotion research. Many emotion researchers have actively suggested \textit{core emotions} or \textit{emotion taxonomy} from their own disciplines, such as evolution, neural system, facial expression, physiology, culture (e.g., \citealp{osgood1966dimensionality, izard1977differential, izard1992basic, plutchik1980general, willcox1982feeling, mano1993assessing, lee2002measuring, cowen2017self, keltner2019emotional}), and language (e.g., \citealp{shaver1987emotion, storm1987taxonomic, hupka1999universal, cowen2019mapping}). The notable points that the studies imply in common are: \textbf{i)} The fixed dimensionality of emotion may not exist, which varies depending on research setting, and \textbf{ii)} The emotion is a complex structure. More than six or seven basic emotions can stand alone. Accordingly, the emotion taxonomy of this study considers the two implications.

We briefly looked at how emotion researchers have constructed the concepts of emotion via emotion vocabulary. One can see that most studies relied on human participants. However, due to the recent advancement of machine learning in natural language processing, words, including emotion words of course, are becoming a full-fledged subject of machine learning. Machine learning methods have introduced many useful tools to obtain rich information of words, which are competent when compared with the traditional approaches in a couple of ways. They are more efficient than the human annotation, and thus allow to handle big language data. They also impose more abundant information on language while the language annotated by a human contains information restricted in a specific research design.

Therefore, in this study, we actively utilize machine learning techniques to follow the fundamental procedure above; identifying and vectorizing emotion words to propose a new emotion structure for the Korean language.

\subsection{Emotion Text Datasets}

In the past few years, many emotion text datasets have been developed, driven by a great interest in emotion analysis. \textbf{Table~\ref{tab:2}} lists currently available Korean emotion text datasets by chronological order of the publication dates.

The datasets are mostly small in size and have rough emotion taxonomies. The lack of a proper emotion corpus is the major motivation of this study.

\section{Korean Emotion Taxonomy}

In this study, we construct a new Korean emotion taxonomy with which our dataset is labeled. The taxonomy is constructed by finding and interpreting the meaning of clusters of emotion concepts. The basic process is as follows: \textbf{i)} Identifying emotion words out of all existing words; \textbf{ii)} Inputting the emotion words into a large pretrained word vector model to obtain a vector for every word; and \textbf{iii)} Clustering the words and interpreting the meaning of the clusters. One interpretable cluster is considered as one emotion in the emotion taxonomy.

\subsection{Emotion Words}

There are a few available emotion words datasets such as Korean Emotion Words Inventory \citep{park2005making}, Korean Emotion Vocabulary Taxonomy \citep{sohn2012korean}, and KNU SentiLex \citep{park2018knu}. KNU SentiLex contains the greatest number of emotion expressions. The researchers preliminarily filtered emotion expressions out of the whole contents of the Korean dictionary by reading the glosses using \texttt{Bi-LSTM} (Bidirectional Long-Short Term Memory; \citealp{hochreiter1997long, schuster1997bidirectional, graves2005framewise}), and manually added emotional slangs and emoticons. Subsequently, they confirmed the emotionality of the expressions by the scrutiny of human raters. As a result, 14k emotion expressions were confirmed and suggested. This study used these three datasets to categorize emotions.

However, the lexicons include some expressions that express emotions figuratively (e.g., \textit{many}). These expressions are excluded because they are more often not used as emotional usage. Moreover, some expressions are missing, and thus we manually added extra expressions. Then, the expressions were tokenized by python package, {\fontfamily{cmss}\selectfont KoNLPy} \citep{park2014konlpy} and function words as well as stop words were deleted. We chose 3,017 expressions that we considered directly represent human emotions, which were inputted into the pretrained word vector model in the next step.

\subsection{Word Vectorization}

The 3,017 emotion words were inputted into a \texttt{fastText} model \citep{bojanowski-etal-2017-enriching} pretrained with large language datasets such as the Korean Wikipedia\footnote{\url{https://github.com/ratsgo/embedding/releases}}. 1,787 words were included in our candidate emotion words list and the model. Hence, the vectors of 1,787 emotion words were used for clustering.

\subsection{Exploring Dimensionality of Emotion}

\textbf{Base Clustering.} The purpose of the \textit{base clustering} is to find the most likely number of clusters of the Korean emotion concepts. In other words, we attempt to answer the question, ‘How many emotions are there, especially in Korean?’ in this stage.

The base clustering is conducted in two steps: \textbf{i)} dimension reduction with \texttt{UMAP} (Uniform Manifold Approximation and Projection; \citealp{mcinnes2018umap}) is performed and \textbf{ii)} the reduced vectors are clustered using \texttt{HDBSCAN} (Hierarchical Density-Based Spatial Clustering of Application with Noise; \citealp{mcinnes2017hdbscan}). The \texttt{HDBSCAN} determines the number of clusters by a survival algorithm. Clusters in a model diminish as its criteria, by which a data point is considered to belong to a cluster, gradually becomes strict and an increasing number of data points are reckoned as noise. Clusters are considered valid, only if they survive long enough in this process. The \texttt{HDBSCAN} estimates the likely number of clusters by this algorithm. Consequently, the number of clusters is given as the final output after the two-step procedure.

\begin{figure*}[p]
\centering
\includegraphics[width=15cm]{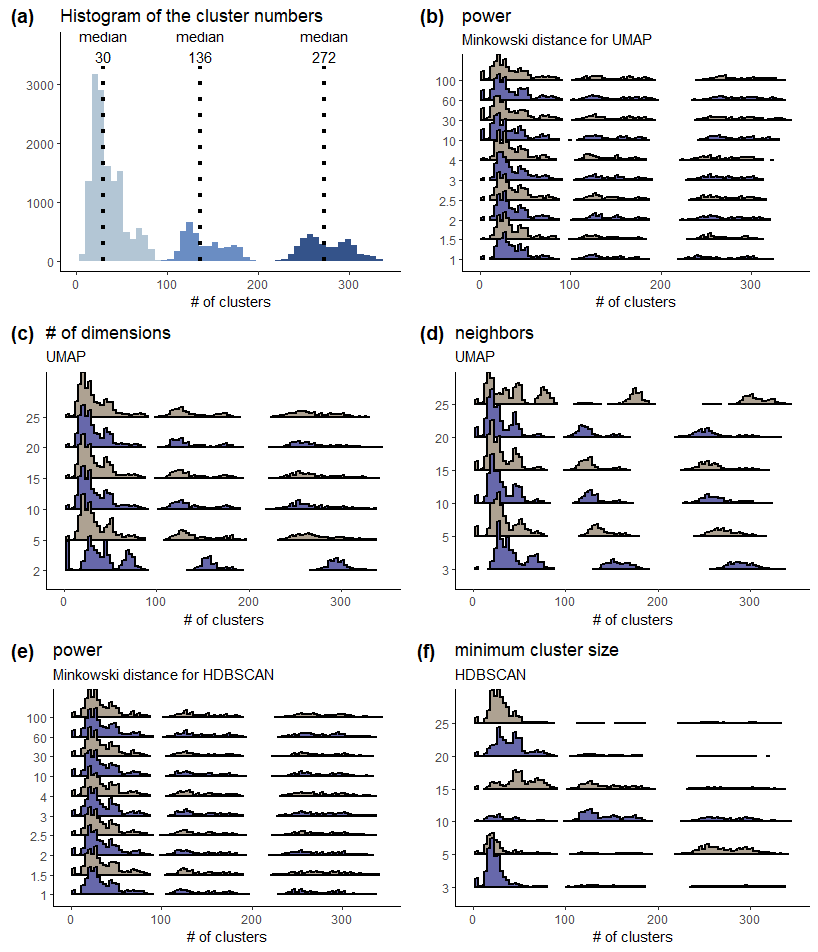}

\caption{
\raggedright{(a) is the histogram for the number of clusters in 21,562 partition sets. Three distributions are identified. (b) – (f) are histograms marginalized on each hyperparameter space. The y-axes represent the searched values of the hyperparameters. Three distributions are consistently identified. The hyperparameters and the number of clusters are not correlated, except for the minimum cluster size (r = -0.2).
(plot packages; {\fontfamily{cmss}\selectfont ggplot2} \citep{wickham2011ggplot2}, {\fontfamily{cmss}\selectfont ggpubr} \citep{kassambara2020package} and {\fontfamily{cmss}\selectfont ggridges} \citep{wilke2021ridgeline}.)
\textbf{Hyperparameters:}
(b): the power in Minkowski distance used to compute the distance matrix for \texttt{UMAP}.
(c): the number of dimensions after the reduction by \texttt{UMAP}.
(d): the number of neighbors of each data point in \texttt{UMAP}.
(e): the power in Minkowski distance used to compute the distance matrix for \texttt{HDBSCAN}.
(f): the minimum size of a group of data points that would be considered as a cluster in \texttt{HDBSCAN}.
}
}
\label{fig:1}
\end{figure*}

The major goal of the two-step strategy is to explore the dimensionality of the emotions as exhaustively as possible. Thus, a grid search was applied on the hyperparameters of each step. The hyperparameters to be searched and the searched values are presented in \textbf{Figure~\ref{fig:1}}. 21,600 points in the hyperparameter space were searched in total.

21,562 partition sets remained, after partition sets with less than three clusters were eliminated. \textbf{Figure~\ref{fig:1}} (a) shows the histogram of the number of clusters. \textbf{Figure~\ref{fig:1}} (b) - (f) show histograms marginalized on each hyperparameter space. Three distributions are robustly identified regardless of the hyperparameters, and the cluster numbers are not correlated to the hyperparameters except for the minimum cluster size. The most likely number of clusters is 30 as in \textbf{Figure~\ref{fig:1}} (a), the median of the largest distribution. This result is consistent with many previous studies. However, we believe that the emotion is so complicated that just 30 categories are insufficient to represent the structure effectively. In addition, recently developed language models are powerful enough to handle complicatedly labeled data. Hence, we decided to proceed for the next most likely number, 136.

\textbf{Clustering Ensemble to Build a New Emotion Taxonomy.} It is not necessary to implement a cluster analysis from scratch to extract 136 clusters, because 21,562 partition sets are already acquired in the base clustering. A cluster ensemble is employed to utilize the partition sets.

The cluster ensemble, literally, is a method that aggregates multiple clustering results to derive one single agreed outcome. We use \texttt{HBGF} (Hybrid Bipartite Graph Formulation; \citealp{fern2004solving}), which exploits both instance- and cluster-based graph formulation (See also \citealp{vega2011survey, karypis1998fast}). In other words, the 21,562 partitions sets were fitted by a \texttt{HBGF} model to reach a consensus of how to split 1,787 emotion words into 136 groups.

The meaning of each cluster is interpreted. Some clusters are considered noninterpretable and dropped because seemingly unrelated words are entangled together. If antonyms are in the same cluster, they are regarded as two separate emotions (i.e., \textit{sadness} and \textit{joy}). 43 emotions were clearly interpreted (see \textbf{Appendix~\ref{sec:appendix_A}}).

\section{KOTE}

We developed KOTE (Korean Online That-gul Emotions), a Korean language dataset containing 50k online comments labeled for the 43 emotions in the new taxonomy. In this chapter, we explain how KOTE is compiled and provide the results of finetuning on a pretrained language model.

\subsection{Text}

50k online comments in KOTE are collected from 12 different platforms of various domains (\textit{news, online community, social media, e-commerce, video platform, movie review, microblog, and forum}) to cover general online environments. The \texttt{robots.txt} guideline of every website was obeyed during the crawling unless no guideline was provided. If a website supports a search engine, randomly selected emotion words from KNU SentiLex were searched for crawling to maximize the emotionality of the collected texts. 3.2 million comments were collected in total, and 50k were sampled being balanced in the number of comments of each website. In the sampling, the minimum length of the texts is set as 10, and the maximum as the 90th percentile of each platform. The grand maximum length is 404, the mean is 57.32, and the median is 42\footnote{The unit of length is a syllable. In the Korean system, 2-3 letters are combined to create one character, which basically corresponds to one syllable. Therefore, the length is 2-3 times longer if the unit is a letter.}.

In all texts, personal information, such as user ID, was deleted without leaving the original. The comments were also supervised for a privacy check by a credible third-party institution designated by the Korea Data Agency, the supporter of this study. They confirmed that no comment contains inappropriate personal information.

\subsection{Label}

The 50k comments were labeled by crowdsourcing in which 3,084 raters whose mother tongue is Korean participated with monetary reward. The labeling procedure is as follows: 50 randomly selected comments are given to a rater. The rater chooses all emotions that the speaker of each comment intends to express. If they identify no emotion, they choose no label but a special label, NO EMOTION. They are also instructed to select plausible emotions and not NO EMOTION, if they think a comment obviously contains some emotion but the exact emotion is not in the given category. Lastly, they are instructed to choose all possibly relevant emotions if the text could have different emotions according to context. The minimum and the maximum number of labels they can choose for one comment are 1 and 10, respectively. The rater can request one more set of 50 comments, and one rater can answer a maximum of two sets. After the labeling, the annotated texts are sent to other crowdworkers who examine the validity of the labels. If the examiner finds labels that they do not agree upon, the disagreed texts are sent back to the original labelers for relabeling. This back-and-forth examination can be repeated three times at maximum.

Two types of catch trials are given in the middle of the labeling. The raters were informed about the catch trials before answering and agreed that the labeling procedure would end with no reward if they did not answer the catch trials correctly. Type-1 catch trial directly instructs the raters to select a certain label, for example \textit{“Please choose only ‘expectancy’ and no other labels for this question”}. Type-2 catch trial asks a question that has a correct answer, for example \textit{“I finally realize what happened. Now I know… I understand everything”}. The selected labels must include \textit{‘realization’}, or the answer is regarded wrong. The correct answer label word is always in the presented text itself.

Five randomly selected raters are assigned to one comment, and thus 250k cases of 50k comments are created as a result. Five binary labels of a comment are summed to be the final label. Thus, the range of a label is 0–5. (see \textbf{Table~\ref{tab:1}}. Three out of the five raters agreed that the text contains \textit{attracted}, so the value of \textit{attracted} label is 3)

\subsection{Data Description}

\textbf{Table~\ref{tab:3}} describes the labels. 99\% of the texts have at least one label of 2 or higher, which means that 99\% have at least one label that two or more raters choose in common. It is evident that the raters did not have much difficulty to reach a consensus. Also, a moderate number of texts are labeled for NO EMOTION.

\begin{table*}[!htbp]
\centering
\begin{tabular}{ccccccc}
\hline
\multicolumn{7}{c}{\textbf{agreement}}                                      \\ \hline
{at least one label of x or higher} & x=1 & x=2 & x=3 & x=4 & x=5 &   \\ \hline
{\begin{tabular}[c]{@{}c@{}}\# of texts\\ \small (\% to total)\end{tabular}} &
  \begin{tabular}[c]{@{}c@{}}50,000\\ \small (100\%)\end{tabular} &
  \begin{tabular}[c]{@{}c@{}}49,663\\ \small (99\%)\end{tabular} &
  \begin{tabular}[c]{@{}c@{}}42,845\\ \small (86\%)\end{tabular} &
  \begin{tabular}[c]{@{}c@{}}28,650\\ \small (57\%)\end{tabular} &
  \begin{tabular}[c]{@{}c@{}}11,760\\ \small (24\%)\end{tabular} &
   \\ \hline
\multicolumn{7}{c}{\textbf{texts labeled for NO EMOTION}}                   \\ \hline
{NO EMOTION}                       & 0   & 1   & 2   & 3   & 4   & 5 \\ \hline
{\begin{tabular}[c]{@{}c@{}}\# of texts\\ \small (\% to total)\end{tabular}} &
  \begin{tabular}[c]{@{}c@{}}42,156\\ \small (84\%)\end{tabular} &
  \begin{tabular}[c]{@{}c@{}}5,243\\ \small (10\%)\end{tabular} &
  \begin{tabular}[c]{@{}c@{}}1,592\\ \small (3\%)\end{tabular} &
  \begin{tabular}[c]{@{}c@{}}644\\ \small (1\%)\end{tabular} &
  \begin{tabular}[c]{@{}c@{}}264\\ \small (0.5\%)\end{tabular} &
  \begin{tabular}[c]{@{}c@{}}101\\ \small (0.2\%)\end{tabular} \\ \hline
\end{tabular}

\caption{\raggedright Description of the labels.}
\label{tab:3}
\end{table*}

\begin{figure*}
\centering
\includegraphics[width=13cm]{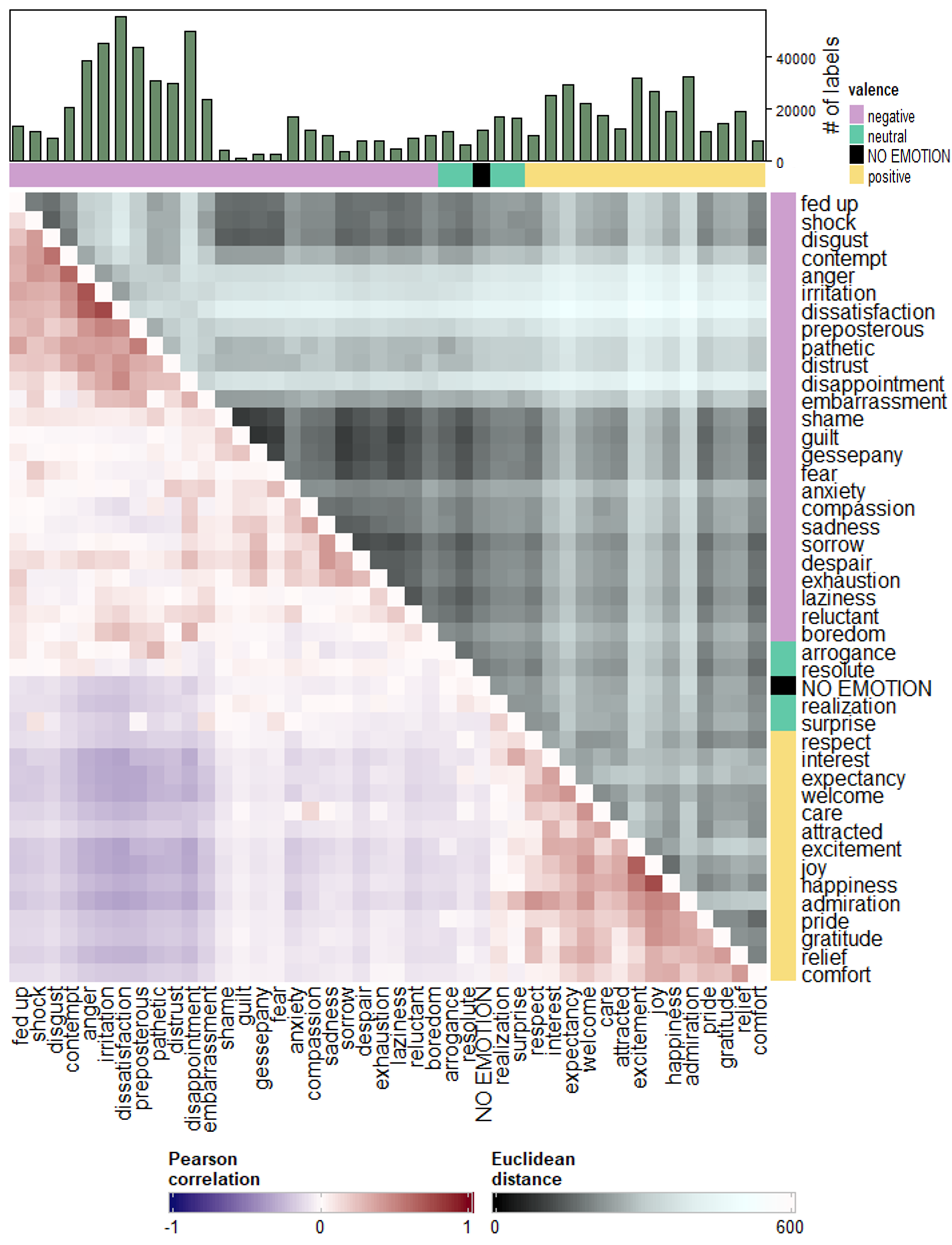}

\caption{\raggedright Heatmap of Pearson correlation and Euclidean distance among the labels. The lower and upper triangle represents the correlation coefficients and the Euclidean distances, respectively. The bars indicate the number of labels in 250k cases. The order of the labels follows Ward clustering with squared Euclidean distance \cite{ward1963hierarchical}. (plot package; {\fontfamily{cmss}\selectfont ComplexHeatmap} \citep{gu2016complex}.)
}
\label{fig:2}
\end{figure*}

The relations among the labels are presented in the heatmap in \textbf{Figure~\ref{fig:2}}. It shows Pearson correlation and Euclidean distance among the labels, each of which is a 50k-dimensional vector.

No additory preprocessing is applied on the data to merge or exclude emotions even though some emotions are linearly related. This is not only because the emotion taxonomy is derived by a nonlinear method, but also the ELECTRA model, which would be finetuned, is nonlinear and potentially able to distinguish linearly similar emotions. In addition, significant emotions differ depending on the method and the criterion. There is no panacea to the best of our knowledge. Lastly, nonsignificant dimensions can additionally provide useful information, despite the risk of redundancy.

\subsection{Finetuning}

\textbf{Preparation.} The labels ranging from 0 to 5 are dichotomized into 0 or 1. Minmax scaling is applied on the labels for each comment. The purpose of the comment-wise minmax scaling is to have the finetuned machine return several possible emotions when no emotion is confidently recognized. The labels exceeding 0.2 after the scaling are converted into 1, and 0 otherwise. One comment has 7.91 labels in average as a result. The dataset is randomly split into train (80\%), test (10\%), and validation (10\%) sets.

\textbf{Training.} We finetuned \texttt{KcELECTRA}, a language model pretrained with Korean online comments, with three packages: {\fontfamily{cmss}\selectfont pytorch} \citep{paszke2019pytorch}, {\fontfamily{cmss}\selectfont pytorch-lightning} \citep{falcon2020framework}, and {\fontfamily{cmss}\selectfont transformers} \citep{wolf2019huggingface}. The batch size is 32, and the input token size is 512. If the number of tokens of an input is less than 512, it is padded with a special token, [PAD]. No input exceeds 512 in length. One linear layer is added on the [CLS] token of the last hidden layer for multi-label classification. The loss is binary cross entropy for each label. We use a linear optimization scheduler, in which the initial learning rate is 2e-5 and the number of warmup steps and total steps are 2,500 and 12,500, respectively. We also switch 5\% of tokens with a random token (except [CLS], [SEP], and [PAD]), and mask 5\% of tokens with a special token, [MASK]. The maximum number of epochs is set as 15, but 9 epochs are enough to reach the optimum in almost all cases. The loss of the validation set is monitored during the learning. We tried label smoothing \citep{szegedy2016rethinking}, but the results are not reported since the performance rather declined. 

\textbf{Results.} The decision threshold for predicted labels is set as 0.3. We use {\fontfamily{cmss}\selectfont scikit-learn} \citep{pedregosa2011scikit} to compute the performance metrics. The average F1-score, AUC (Area Under Curve; \citealp{hanley1982meaning}), and MCC (Mathews Correlation Coefficient; \citealp{matthews1975comparison, baldi2000assessing, chicco2020advantages}) are 0.56, 0.88, and 0.59, respectively (see \textbf{Appendix~\ref{sec:appendix_B}} for full description).

As mentioned in the Introduction section, these results are obtained with arbitrarily decided hyperparameters. Therefore, the performance can be improved with additional methods, such as hyperparameter tuning. Otherwise, it would be a good attempt to employ different approaches for the preprocessing, such as label merging, dichotomization, or label balancing. Since the dataset is fully open, one can try anything necessary. If a good result is obtained, we hope it would be shared without hesitation.

\section{Conclusions}

The model finetuned with our dataset achieved a better performance than the existing model finetuned with the translated GoEmotions dataset (F1-scores are 0.56 versus 0.41). Although direct comparison is difficult because of different emotion taxonomies, it is meaningful to achieve a comparable performance with a wider range of emotions (43 emotions versus 27 emotions). The reasons for good performance can be summarized as follows. \textbf{i)} We derived emotion taxonomy by introducing machine learning to repeatedly validated psychological theories and methodologies. \textbf{ii)} The emotion taxonomy is befitting to Korean culture, which is beneficial in two respects; the human raters can easily understand the emotions in the taxonomy, and the Korean language model can infer the emotions of the texts efficiently. \textbf{iii)} We viewed the emotion as a complex structure according to the existing psychology literature, which motivated us to impose complex information on the texts in the labeling and to maintain the complexity in the preprocessing.

\section{Limitations}

However, there are limitations that the users should keep in mind:
\textbf{i)}	Emotion is a complex structure, which is impossible to perfectly capture with just tens of emotions.
\textbf{ii)} Emotion is a dynamic structure, but we treat it as a static structure in this study. The emotions must interact complicatedly. For example, an emotion may be combined with other emotions to create a new one, or one single emotion can have different meanings according to the degree of emotionality and contextuality.
\textbf{iii)} KOTE is large, but not large enough to cover different domains inside and outside the internet. KOTE may have limitations when one tries to apply the trained model to a different type of texts other than online comments. \textit{Fear}, for example, is one of the core emotions but rarely appears in our dataset. Accordingly, linguistic expressions associated with \textit{fear} might be scarce as well.
\textbf{iv)} The discriminatory evaluation against protected groups is carried within our dataset, since it reflects the discrimination of the texts and the human raters. We highly recommend \textbf{Appendix~\ref{sec:appendix_C}} for ethical consideration.

Although future works are required to answer those questions, KOTE is still a new useful tool that helps to overstep the limit of mere sentiment analysis. We hope this user guide provides the users with useful information to utilize the dataset.

\section*{Acknowledgements}

This study is supported by the 2021 Data Voucher Support Project organized by the Korea Data Agency under the Ministry of Science and ICT of the Government of South Korea. We also thank all crowdworkers who sincerely helped us to annotate the data.

\bibliography{KOTE}

\begin{thebibliography}{59}
\expandafter\ifx\csname natexlab\endcsname\relax\def\natexlab#1{#1}\fi

\bibitem[{AIHUB(2021)}]{aihub2021dialogue}
AIHUB. 2021.
\newblock Emotional dialogue corpus.
\newblock \url{https://aihub.or.kr/aidata/7978}.

\bibitem[{Averill(1975)}]{averill1975semantic}
James~R Averill. 1975.
\newblock \emph{A semantic atlas of emotional concepts}.
\newblock American Psycholog. Ass., Journal Suppl. Abstract Service.

\bibitem[{Baldi et~al.(2000)Baldi, Brunak, Chauvin, Andersen, and
  Nielsen}]{baldi2000assessing}
Pierre Baldi, S{\o}ren Brunak, Yves Chauvin, Claus~AF Andersen, and Henrik
  Nielsen. 2000.
\newblock Assessing the accuracy of prediction algorithms for classification:
  an overview.
\newblock \emph{Bioinformatics}, 16(5):412--424.

\bibitem[{Block(1957)}]{block1957studies}
Jack Block. 1957.
\newblock Studies in the phenomenology of emotions.
\newblock \emph{The Journal of Abnormal and Social Psychology}, 54(3):358.

\bibitem[{Bojanowski et~al.(2017)Bojanowski, Grave, Joulin, and
  Mikolov}]{bojanowski-etal-2017-enriching}
Piotr Bojanowski, Edouard Grave, Armand Joulin, and Tomas Mikolov. 2017.
\newblock \href {https://doi.org/10.1162/tacl_a_00051} {Enriching word vectors
  with subword information}.
\newblock \emph{Transactions of the Association for Computational Linguistics},
  5:135--146.

\bibitem[{Caton and Haas(2020)}]{caton2020fairness}
Simon Caton and Christian Haas. 2020.
\newblock Fairness in machine learning: A survey.
\newblock \emph{arXiv preprint arXiv:2010.04053}.

\bibitem[{Chicco and Jurman(2020)}]{chicco2020advantages}
Davide Chicco and Giuseppe Jurman. 2020.
\newblock The advantages of the matthews correlation coefficient (mcc) over f1
  score and accuracy in binary classification evaluation.
\newblock \emph{BMC genomics}, 21(1):1--13.

\bibitem[{Clark et~al.(2020)Clark, Luong, Le, and Manning}]{clark2020electra}
Kevin Clark, Minh-Thang Luong, Quoc~V Le, and Christopher~D Manning. 2020.
\newblock Electra: Pre-training text encoders as discriminators rather than
  generators.
\newblock \emph{arXiv preprint arXiv:2003.10555}.

\bibitem[{Clore et~al.(1987)Clore, Ortony, and Foss}]{clore1987psychological}
Gerald~L Clore, Andrew Ortony, and Mark~A Foss. 1987.
\newblock The psychological foundations of the affective lexicon.
\newblock \emph{Journal of personality and social psychology}, 53(4):751.

\bibitem[{Cowen et~al.(2019)Cowen, Sauter, Tracy, and
  Keltner}]{cowen2019mapping}
Alan Cowen, Disa Sauter, Jessica~L Tracy, and Dacher Keltner. 2019.
\newblock Mapping the passions: Toward a high-dimensional taxonomy of emotional
  experience and expression.
\newblock \emph{Psychological Science in the Public Interest}, 20(1):69--90.

\bibitem[{Cowen and Keltner(2017)}]{cowen2017self}
Alan~S Cowen and Dacher Keltner. 2017.
\newblock Self-report captures 27 distinct categories of emotion bridged by
  continuous gradients.
\newblock \emph{Proceedings of the National Academy of Sciences},
  114(38):E7900--E7909.

\bibitem[{Demszky et~al.(2020)Demszky, Movshovitz-Attias, Ko, Cowen, Nemade,
  and Ravi}]{demszky-etal-2020-goemotions}
Dorottya Demszky, Dana Movshovitz-Attias, Jeongwoo Ko, Alan Cowen, Gaurav
  Nemade, and Sujith Ravi. 2020.
\newblock \href {https://doi.org/10.18653/v1/2020.acl-main.372}
  {{G}o{E}motions: A dataset of fine-grained emotions}.
\newblock In \emph{Proceedings of the 58th Annual Meeting of the Association
  for Computational Linguistics}, pages 4040--4054, Online. Association for
  Computational Linguistics.

\bibitem[{Falcon and Cho(2020)}]{falcon2020framework}
William Falcon and Kyunghyun Cho. 2020.
\newblock A framework for contrastive self-supervised learning and designing a
  new approach.
\newblock \emph{arXiv preprint arXiv:2009.00104}.

\bibitem[{Fern and Brodley(2004)}]{fern2004solving}
Xiaoli~Zhang Fern and Carla~E Brodley. 2004.
\newblock Solving cluster ensemble problems by bipartite graph partitioning.
\newblock In \emph{Proceedings of the twenty-first international conference on
  Machine learning}, page~36.

\bibitem[{Gendron et~al.(2014)Gendron, Roberson, van~der Vyver, and
  Barrett}]{gendron2014perceptions}
Maria Gendron, Debi Roberson, Jacoba~Marietta van~der Vyver, and Lisa~Feldman
  Barrett. 2014.
\newblock Perceptions of emotion from facial expressions are not culturally
  universal: evidence from a remote culture.
\newblock \emph{Emotion}, 14(2):251.

\bibitem[{Graves and Schmidhuber(2005)}]{graves2005framewise}
Alex Graves and J{\"u}rgen Schmidhuber. 2005.
\newblock Framewise phoneme classification with bidirectional lstm networks.
\newblock In \emph{Proceedings. 2005 IEEE International Joint Conference on
  Neural Networks, 2005.}, volume~4, pages 2047--2052. IEEE.

\bibitem[{Gu et~al.(2016)Gu, Eils, and Schlesner}]{gu2016complex}
Zuguang Gu, Roland Eils, and Matthias Schlesner. 2016.
\newblock Complex heatmaps reveal patterns and correlations in multidimensional
  genomic data.
\newblock \emph{Bioinformatics}, 32(18):2847--2849.

\bibitem[{Hanley and McNeil(1982)}]{hanley1982meaning}
James~A Hanley and Barbara~J McNeil. 1982.
\newblock The meaning and use of the area under a receiver operating
  characteristic (roc) curve.
\newblock \emph{Radiology}, 143(1):29--36.

\bibitem[{Hochreiter and Schmidhuber(1997)}]{hochreiter1997long}
Sepp Hochreiter and J{\"u}rgen Schmidhuber. 1997.
\newblock Long short-term memory.
\newblock \emph{Neural computation}, 9(8):1735--1780.

\bibitem[{Hupka et~al.(1999)Hupka, Lenton, and Hutchison}]{hupka1999universal}
Ralph~B Hupka, Alison~P Lenton, and Keith~A Hutchison. 1999.
\newblock Universal development of emotion categories in natural language.
\newblock \emph{Journal of personality and social psychology}, 77(2):247.

\bibitem[{Izard(1977)}]{izard1977differential}
Carroll~E Izard. 1977.
\newblock Differential emotions theory.
\newblock In \emph{Human emotions}, pages 43--66. Springer.

\bibitem[{Izard(1992)}]{izard1992basic}
Carroll~E Izard. 1992.
\newblock Basic emotions, relations among emotions, and emotion-cognition
  relations.

\bibitem[{Jang et~al.(2013)Jang, Kim, and Shin}]{jang-etal-2013-kosac}
Hayeon Jang, Munhyong Kim, and Hyopil Shin. 2013.
\newblock \href {https://aclanthology.org/Y13-1037} {{KOSAC}: A full-fledged
  {K}orean sentiment analysis corpus}.
\newblock In \emph{Proceedings of the 27th Pacific Asia Conference on Language,
  Information, and Computation ({PACLIC} 27)}, pages 366--373, Taipei, Taiwan.
  Department of English, National Chengchi University.

\bibitem[{Karypis and Kumar(1998)}]{karypis1998fast}
George Karypis and Vipin Kumar. 1998.
\newblock A fast and high quality multilevel scheme for partitioning irregular
  graphs.
\newblock \emph{SIAM Journal on scientific Computing}, 20(1):359--392.

\bibitem[{Kassambara and Kassambara(2020)}]{kassambara2020package}
Alboukadel Kassambara and Maintainer~Alboukadel Kassambara. 2020.
\newblock Package ‘ggpubr’.
\newblock \emph{R package version 0.1}, 6.

\bibitem[{Keltner et~al.(2019)Keltner, Sauter, Tracy, and
  Cowen}]{keltner2019emotional}
Dacher Keltner, Disa Sauter, Jessica Tracy, and Alan Cowen. 2019.
\newblock Emotional expression: Advances in basic emotion theory.
\newblock \emph{Journal of nonverbal behavior}, 43(2):133--160.

\bibitem[{KETI(2020{\natexlab{a}})}]{keti2020continuous}
KETI. 2020{\natexlab{a}}.
\newblock Korean continuous dialogue dataset with emotion information.
\newblock
  \url{https://aihub.or.kr/opendata/keti-data/recognition-laguage/KETI-02-010}.

\bibitem[{KETI(2020{\natexlab{b}})}]{keti2020oneoff}
KETI. 2020{\natexlab{b}}.
\newblock Korean one-off dialogue dataset with emotion information.
\newblock
  \url{https://aihub.or.kr/opendata/keti-data/recognition-laguage/KETI-02-009}.

\bibitem[{Lee and Lim(2002)}]{lee2002measuring}
Hak-Sik Lee and Ji~Hoon Lim. 2002.
\newblock Measuring the consumption-related emotion construct.
\newblock \emph{Korea Marketing Review}, 17(3):55--91.

\bibitem[{Lee(2021)}]{lee2021kcelectra}
Junbum Lee. 2021.
\newblock Kcelectra: Korean comments electra.
\newblock \url{https://github.com/Beomi/KcELECTRA}.

\bibitem[{Mano and Oliver(1993)}]{mano1993assessing}
Haim Mano and Richard~L Oliver. 1993.
\newblock Assessing the dimensionality and structure of the consumption
  experience: evaluation, feeling, and satisfaction.
\newblock \emph{Journal of Consumer research}, 20(3):451--466.

\bibitem[{Matthews(1975)}]{matthews1975comparison}
Brian~W Matthews. 1975.
\newblock Comparison of the predicted and observed secondary structure of t4
  phage lysozyme.
\newblock \emph{Biochimica et Biophysica Acta (BBA)-Protein Structure},
  405(2):442--451.

\bibitem[{McInnes et~al.(2017)McInnes, Healy, and Astels}]{mcinnes2017hdbscan}
Leland McInnes, John Healy, and Steve Astels. 2017.
\newblock hdbscan: Hierarchical density based clustering.
\newblock \emph{J. Open Source Softw.}, 2(11):205.

\bibitem[{McInnes et~al.(2018)McInnes, Healy, and Melville}]{mcinnes2018umap}
Leland McInnes, John Healy, and James Melville. 2018.
\newblock Umap: Uniform manifold approximation and projection for dimension
  reduction.
\newblock \emph{arXiv preprint arXiv:1802.03426}.

\bibitem[{Mehrabi et~al.(2021)Mehrabi, Morstatter, Saxena, Lerman, and
  Galstyan}]{mehrabi2021survey}
Ninareh Mehrabi, Fred Morstatter, Nripsuta Saxena, Kristina Lerman, and Aram
  Galstyan. 2021.
\newblock A survey on bias and fairness in machine learning.
\newblock \emph{ACM Computing Surveys (CSUR)}, 54(6):1--35.

\bibitem[{Mesquita and Frijda(1992)}]{mesquita1992cultural}
Batja Mesquita and Nico~H Frijda. 1992.
\newblock Cultural variations in emotions: a review.
\newblock \emph{Psychological bulletin}, 112(2):179.

\bibitem[{Naver(2015)}]{nsmc2015}
Naver. 2015.
\newblock Nsmc: Naver sentiment movie corpus.
\newblock \url{https://github.com/e9t/nsmc}.

\bibitem[{Osgood(1966)}]{osgood1966dimensionality}
Charles~E Osgood. 1966.
\newblock Dimensionality of the semantic space for communication via facial
  expressions.
\newblock \emph{Scandinavian journal of psychology}, 7(1):1--30.

\bibitem[{Park and Cho(2014)}]{park2014konlpy}
E~Park and S~Cho. 2014.
\newblock Konlpy: easy and concise korean information processing python
  package.
\newblock In \emph{Proceedings of the 26th Korean and Korean Information
  Processing Conference}, pages 1--4.

\bibitem[{Park and Min(2005)}]{park2005making}
In-Jo Park and Kyung-Hwan Min. 2005.
\newblock Making a list of korean emotion terms and exploring dimensions
  underlying them.
\newblock \emph{Korean Journal of Social and Personality Psychology},
  19(1):109--129.

\bibitem[{Park et~al.(2018{\natexlab{a}})Park, Shin, and
  Fung}]{park-etal-2018-reducing}
Ji~Ho Park, Jamin Shin, and Pascale Fung. 2018{\natexlab{a}}.
\newblock \href {https://doi.org/10.18653/v1/D18-1302} {Reducing gender bias in
  abusive language detection}.
\newblock In \emph{Proceedings of the 2018 Conference on Empirical Methods in
  Natural Language Processing}, pages 2799--2804, Brussels, Belgium.
  Association for Computational Linguistics.

\bibitem[{Park et~al.(2018{\natexlab{b}})Park, Na, Choi, Lee, and
  On}]{park2018knu}
Sang-Min Park, Chul-Won Na, Min-Seong Choi, Da-Hee Lee, and Byung-Won On.
  2018{\natexlab{b}}.
\newblock Knu korean sentiment lexicon: Bi-lstm-based method for building a
  korean sentiment lexicon.
\newblock \emph{Journal of Intelligence and Information Systems},
  24(4):219--240.

\bibitem[{Paszke et~al.(2019)Paszke, Gross, Massa, Lerer, Bradbury, Chanan,
  Killeen, Lin, Gimelshein, Antiga et~al.}]{paszke2019pytorch}
Adam Paszke, Sam Gross, Francisco Massa, Adam Lerer, James Bradbury, Gregory
  Chanan, Trevor Killeen, Zeming Lin, Natalia Gimelshein, Luca Antiga, et~al.
  2019.
\newblock Pytorch: An imperative style, high-performance deep learning library.
\newblock \emph{Advances in neural information processing systems}, 32.

\bibitem[{Pedregosa et~al.(2011)Pedregosa, Varoquaux, Gramfort, Michel,
  Thirion, Grisel, Blondel, Prettenhofer, Weiss, Dubourg
  et~al.}]{pedregosa2011scikit}
Fabian Pedregosa, Ga{\"e}l Varoquaux, Alexandre Gramfort, Vincent Michel,
  Bertrand Thirion, Olivier Grisel, Mathieu Blondel, Peter Prettenhofer, Ron
  Weiss, Vincent Dubourg, et~al. 2011.
\newblock Scikit-learn: Machine learning in python.
\newblock \emph{the Journal of machine Learning research}, 12:2825--2830.

\bibitem[{Plutchik(1980)}]{plutchik1980general}
Robert Plutchik. 1980.
\newblock A general psychoevolutionary theory of emotion.
\newblock In \emph{Theories of emotion}, pages 3--33. Elsevier.

\bibitem[{Schuster and Paliwal(1997)}]{schuster1997bidirectional}
Mike Schuster and Kuldip~K Paliwal. 1997.
\newblock Bidirectional recurrent neural networks.
\newblock \emph{IEEE transactions on Signal Processing}, 45(11):2673--2681.

\bibitem[{Shaver et~al.(1987)Shaver, Schwartz, Kirson, and
  O'connor}]{shaver1987emotion}
Phillip Shaver, Judith Schwartz, Donald Kirson, and Cary O'connor. 1987.
\newblock Emotion knowledge: further exploration of a prototype approach.
\newblock \emph{Journal of personality and social psychology}, 52(6):1061.

\bibitem[{Shields(1984)}]{shields1984distinguishing}
Stephanie~A Shields. 1984.
\newblock Distinguishing between emotion and nonemotion: Judgments about
  experience.
\newblock \emph{Motivation and Emotion}, 8(4):355--369.

\bibitem[{Sohn et~al.(2012)Sohn, Park, Park, and Sohn}]{sohn2012korean}
Sun-Ju Sohn, Mi-Sook Park, Ji-Eun Park, and Jin-Hun Sohn. 2012.
\newblock Korean emotion vocabulary: extraction and categorization of feeling
  words.
\newblock \emph{Science of Emotion and Sensibility}, 15(1):105--120.

\bibitem[{Storm and Storm(1987)}]{storm1987taxonomic}
Christine Storm and Tom Storm. 1987.
\newblock A taxonomic study of the vocabulary of emotions.
\newblock \emph{Journal of personality and social psychology}, 53(4):805.

\bibitem[{Sun et~al.(2019)Sun, Gaut, Tang, Huang, ElSherief, Zhao, Mirza,
  Belding, Chang, and Wang}]{sun-etal-2019-mitigating}
Tony Sun, Andrew Gaut, Shirlyn Tang, Yuxin Huang, Mai ElSherief, Jieyu Zhao,
  Diba Mirza, Elizabeth Belding, Kai-Wei Chang, and William~Yang Wang. 2019.
\newblock \href {https://doi.org/10.18653/v1/P19-1159} {Mitigating gender bias
  in natural language processing: Literature review}.
\newblock In \emph{Proceedings of the 57th Annual Meeting of the Association
  for Computational Linguistics}, pages 1630--1640, Florence, Italy.
  Association for Computational Linguistics.

\bibitem[{Szegedy et~al.(2016)Szegedy, Vanhoucke, Ioffe, Shlens, and
  Wojna}]{szegedy2016rethinking}
Christian Szegedy, Vincent Vanhoucke, Sergey Ioffe, Jon Shlens, and Zbigniew
  Wojna. 2016.
\newblock Rethinking the inception architecture for computer vision.
\newblock In \emph{Proceedings of the IEEE conference on computer vision and
  pattern recognition}, pages 2818--2826.

\bibitem[{Vega-Pons and Ruiz-Shulcloper(2011)}]{vega2011survey}
Sandro Vega-Pons and Jos{\'e} Ruiz-Shulcloper. 2011.
\newblock A survey of clustering ensemble algorithms.
\newblock \emph{International Journal of Pattern Recognition and Artificial
  Intelligence}, 25(03):337--372.

\bibitem[{Ward~Jr(1963)}]{ward1963hierarchical}
Joe~H Ward~Jr. 1963.
\newblock Hierarchical grouping to optimize an objective function.
\newblock \emph{Journal of the American statistical association},
  58(301):236--244.

\bibitem[{Wickham(2011)}]{wickham2011ggplot2}
Hadley Wickham. 2011.
\newblock ggplot2.
\newblock \emph{Wiley interdisciplinary reviews: computational statistics},
  3(2):180--185.

\bibitem[{Wilke(2021)}]{wilke2021ridgeline}
Claus~O Wilke. 2021.
\newblock Ridgeline plots in ‘ggplot2’[r package ggridges version 0.5. 3].
\newblock \emph{January. https://cran. r-project.
  org/web/packages/ggridges/index. html}.

\bibitem[{Willcox(1982)}]{willcox1982feeling}
Gloria Willcox. 1982.
\newblock The feeling wheel: A tool for expanding awareness of emotions and
  increasing spontaneity and intimacy.
\newblock \emph{Transactional Analysis Journal}, 12(4):274--276.

\bibitem[{Wolf et~al.(2019)Wolf, Debut, Sanh, Chaumond, Delangue, Moi, Cistac,
  Rault, Louf, Funtowicz et~al.}]{wolf2019huggingface}
Thomas Wolf, Lysandre Debut, Victor Sanh, Julien Chaumond, Clement Delangue,
  Anthony Moi, Pierric Cistac, Tim Rault, R{\'e}mi Louf, Morgan Funtowicz,
  et~al. 2019.
\newblock Huggingface's transformers: State-of-the-art natural language
  processing.
\newblock \emph{arXiv preprint arXiv:1910.03771}.

\bibitem[{Zhao et~al.(2018)Zhao, Wang, Yatskar, Ordonez, and
  Chang}]{zhao-etal-2018-gender}
Jieyu Zhao, Tianlu Wang, Mark Yatskar, Vicente Ordonez, and Kai-Wei Chang.
  2018.
\newblock \href {https://doi.org/10.18653/v1/N18-2003} {Gender bias in
  coreference resolution: Evaluation and debiasing methods}.
\newblock In \emph{Proceedings of the 2018 Conference of the North {A}merican
  Chapter of the Association for Computational Linguistics: Human Language
  Technologies, Volume 2 (Short Papers)}, pages 15--20, New Orleans, Louisiana.
  Association for Computational Linguistics.

\end{thebibliography}
\bibliographystyle{acl_natbib}

\appendix

\onecolumn
\section{Appendix: Emotion Clusters}
\label{sec:appendix_A}

\begin{table*}[h!]
\centering
\begin{tabular}{c|c|c}
\hline
\textbf{Valence}                    & \textbf{Interpretation}  & \textbf{Example words in the cluster}                    \\ \hline
\multirow{25}{*}{\textbf{Negative}} & \textit{dissatisfaction} & dissatisfied, oppose, criticize, complaint               \\ \cline{2-3} 
                                    & \textit{embarrassment}   & embarrassed, disconcerted, awkward, untoward             \\ \cline{2-3} 
                                    & \textit{irritation}      & irritated, pissed off, ridiculous                        \\ \cline{2-3} 
                                    & \textit{sadness}         & sad, miss, lonely, tear                                  \\ \cline{2-3} 
                                    & \textit{despair}         & frustrated, joys \& sorrows, hurt, grief, letdown        \\ \cline{2-3} 
                                    & \textit{shame}           & ashamed, humiliated                                      \\ \cline{2-3} 
                                    & \textit{boredom}         & bored, tedium, trite, dull                               \\ \cline{2-3} 
                                    & \textit{disappointment}  & disappointed, sorry, upset, deplorable, regretful        \\ \cline{2-3} 
                                    & \textit{disgust}         & disgusted, repulsive, dirty                              \\ \cline{2-3} 
                                    & \textit{shock}           & shocked, flabbergasted, pass out, freaked out            \\ \cline{2-3} 
                                    & \textit{reluctant}       & unwilling, denial, pressure, cannot be bothered, give up \\ \cline{2-3} 
                                    & \textit{fear}            & fear, anxious, tense, pressed                            \\ \cline{2-3} 
                                    & \textit{contempt}        & contempt, hatred, scorn, vilifying                       \\ \cline{2-3} 
                                    & \textit{guilt}           & guilt, blamed, repentance, remorse                       \\ \cline{2-3} 
                                    & \textit{anxiety}         & apprehensive, worry, threatened                          \\ \cline{2-3} 
                                    & \textit{distrust}        & suspicious, doubtful, lie                                \\ \cline{2-3} 
                                    & \textit{anger}           & anger, rage, obsessed, fury                              \\ \cline{2-3} 
                                    & \textit{gessepany}       & failure, miserably, extorted                             \\ \cline{2-3} 
                                    & \textit{laziness}        & bothered, dawdling                                       \\ \cline{2-3} 
                                    & \textit{sorrow}          & sorrowful, mirthless, weary, sobbing, upset, complicated \\ \cline{2-3} 
                                    & \textit{fed up}          & fed up, struggle, arduous, sick and tired                \\ \cline{2-3} 
                                    & \textit{preposterous}    & dumbfounded, stunned, sttufy, enervated, WTF             \\ \cline{2-3} 
                                    & \textit{compassion}      & pity, sadly, chocked up, heartrending                    \\ \cline{2-3} 
                                    & \textit{pathetic}        & pathetic, belittled, stupid, impudence                   \\ \cline{2-3} 
                                    & \textit{exhaustion}      & tired, peak, exhausted                                   \\ \hline
\multirow{14}{*}{\textbf{Positive}} & \textit{admiration}      & admiring, great, praise, compliment                      \\ \cline{2-3} 
                                    & \textit{happiness}       & happy, affection, valuable, hope, luck                   \\ \cline{2-3} 
                                    & \textit{joy}             & delight, ecstasy, love                                   \\ \cline{2-3} 
                                    & \textit{gratitude}       & praiseworthy, commendable, favor, blessing, mercy        \\ \cline{2-3} 
                                    & \textit{excitement}      & excited, funny                                           \\ \cline{2-3} 
                                    & \textit{care}            & caring, adore, dear                                      \\ \cline{2-3} 
                                    & \textit{expectancy}      & new, achieve, together, harmonious, vitality             \\ \cline{2-3} 
                                    & \textit{comfort}         & comfortable, ease, cozy, cool, warm                      \\ \cline{2-3} 
                                    & \textit{welcome}         & welcome, approval, kindness, enthusiastic                \\ \cline{2-3} 
                                    & \textit{interest}        & interested, curious                                      \\ \cline{2-3} 
                                    & \textit{relief}          & relief, trust, intimate, close                           \\ \cline{2-3} 
                                    & \textit{respect}         & respect, loyal, veneration, follow, obedience            \\ \cline{2-3} 
                                    & \textit{attracted}       & handsome, pretty, sweet, thrilled, cute, aegyo           \\ \cline{2-3} 
                                    & \textit{pride}           & successful, victory, worthwhile, accomplish              \\ \hline
\multirow{4}{*}{\textbf{Neutral}}   & \textit{arrogance}       & arrogance, pompous, ignore, bragging, boast, gasconade   \\ \cline{2-3} 
                                    & \textit{surprise}        & astonished, startled                                     \\ \cline{2-3} 
                                    & \textit{realization}     & realize, enlightened, wakened, conviction, belief        \\ \cline{2-3} 
                                    & \textit{resolute}        & resolute, determination                                  \\ \hline
\end{tabular}

\caption{\raggedright Interpretation of each interpretable cluster and emotion words in it.}
\end{table*}

\clearpage
\section{Appendix: Performance Metrics}
\label{sec:appendix_B}

\begin{table*}[!h]
\resizebox{\textwidth}{!}{\begin{tabular}{lclclclc|lllclclcc}
\hline\hline
\multicolumn{17}{c}{\textbf{F1-score}} \\ \hline
\multicolumn{1}{c}{\textbf{emotion}} &
  \multicolumn{2}{c}{\textbf{precision}} &
  \multicolumn{2}{c}{\textbf{recall}} &
  \multicolumn{2}{c}{\textbf{F1}} &
  \textbf{\#} &
  \multicolumn{3}{c}{\textbf{emotion}} &
  \multicolumn{2}{c}{\textbf{precision}} &
  \multicolumn{2}{c}{\textbf{recall}} &
  \textbf{F1} &
  \textbf{\#} \\ \hline
\textit{dissatisfaction} &
  \multicolumn{2}{c}{0.78} &
  \multicolumn{2}{c}{0.89} &
  \multicolumn{2}{c}{0.83} &
  2113 &
  \multicolumn{3}{l}{\textit{admiration}} &
  \multicolumn{2}{c}{0.67} &
  \multicolumn{2}{c}{0.86} &
  0.75 &
  1323 \\ \hline
\textit{embarrassment} &
  \multicolumn{2}{c}{0.57} &
  \multicolumn{2}{c}{0.70} &
  \multicolumn{2}{c}{0.63} &
  1319 &
  \multicolumn{3}{l}{\textit{happiness}} &
  \multicolumn{2}{c}{0.57} &
  \multicolumn{2}{c}{0.80} &
  0.67 &
  906 \\ \hline
\textit{irritation} &
  \multicolumn{2}{c}{0.74} &
  \multicolumn{2}{c}{0.86} &
  \multicolumn{2}{c}{0.80} &
  1909 &
  \multicolumn{3}{l}{\textit{joy}} &
  \multicolumn{2}{c}{0.65} &
  \multicolumn{2}{c}{0.85} &
  0.73 &
  1205 \\ \hline
\textit{sadness} &
  \multicolumn{2}{c}{0.62} &
  \multicolumn{2}{c}{0.61} &
  \multicolumn{2}{c}{0.62} &
  545 &
  \multicolumn{3}{l}{\textit{gratitude}} &
  \multicolumn{2}{c}{0.54} &
  \multicolumn{2}{c}{0.70} &
  0.61 &
  637 \\ \hline
\textit{despair} &
  \multicolumn{2}{c}{0.46} &
  \multicolumn{2}{c}{0.41} &
  \multicolumn{2}{c}{0.43} &
  472 &
  \multicolumn{3}{l}{\textit{excitement}} &
  \multicolumn{2}{c}{0.69} &
  \multicolumn{2}{c}{0.86} &
  0.77 &
  1321 \\ \hline
\textit{shame} &
  \multicolumn{2}{c}{0.30} &
  \multicolumn{2}{c}{0.05} &
  \multicolumn{2}{c}{0.08} &
  306 &
  \multicolumn{3}{l}{\textit{care}} &
  \multicolumn{2}{c}{0.56} &
  \multicolumn{2}{c}{0.69} &
  0.62 &
  897 \\ \hline
\textit{boredom} &
  \multicolumn{2}{c}{0.67} &
  \multicolumn{2}{c}{0.54} &
  \multicolumn{2}{c}{0.60} &
  470 &
  \multicolumn{3}{l}{\textit{expectancy}} &
  \multicolumn{2}{c}{0.58} &
  \multicolumn{2}{c}{0.81} &
  0.67 &
  1359 \\ \hline
\textit{disappointment} &
  \multicolumn{2}{c}{0.68} &
  \multicolumn{2}{c}{0.88} &
  \multicolumn{2}{c}{0.77} &
  2185 &
  \multicolumn{3}{l}{\textit{comfort}} &
  \multicolumn{2}{c}{0.45} &
  \multicolumn{2}{c}{0.51} &
  0.48 &
  458 \\ \hline
\textit{disgust} &
  \multicolumn{2}{c}{0.48} &
  \multicolumn{2}{c}{0.59} &
  \multicolumn{2}{c}{0.53} &
  516 &
  \multicolumn{3}{l}{\textit{welcome}} &
  \multicolumn{2}{c}{0.56} &
  \multicolumn{2}{c}{0.83} &
  0.67 &
  1109 \\ \hline
\textit{shock} &
  \multicolumn{2}{c}{0.45} &
  \multicolumn{2}{c}{0.50} &
  \multicolumn{2}{c}{0.47} &
  704 &
  \multicolumn{3}{l}{\textit{interest}} &
  \multicolumn{2}{c}{0.57} &
  \multicolumn{2}{c}{0.77} &
  0.66 &
  1346 \\ \hline
\textit{reluctant} &
  \multicolumn{2}{c}{0.43} &
  \multicolumn{2}{c}{0.33} &
  \multicolumn{2}{c}{0.37} &
  606 &
  \multicolumn{3}{l}{\textit{relief}} &
  \multicolumn{2}{c}{0.53} &
  \multicolumn{2}{c}{0.75} &
  0.62 &
  945 \\ \hline
\textit{fear} &
  \multicolumn{2}{c}{0.36} &
  \multicolumn{2}{c}{0.26} &
  \multicolumn{2}{c}{0.30} &
  164 &
  \multicolumn{3}{l}{\textit{respect}} &
  \multicolumn{2}{c}{0.52} &
  \multicolumn{2}{c}{0.68} &
  0.59 &
  460 \\ \hline
\textit{contempt} &
  \multicolumn{2}{c}{0.66} &
  \multicolumn{2}{c}{0.77} &
  \multicolumn{2}{c}{0.71} &
  984 &
  \multicolumn{3}{l}{\textit{attracted}} &
  \multicolumn{2}{c}{0.60} &
  \multicolumn{2}{c}{0.64} &
  0.62 &
  524 \\ \hline
\textit{guilt} &
  \multicolumn{2}{c}{0.00} &
  \multicolumn{2}{c}{0.00} &
  \multicolumn{2}{c}{0.00} &
  84 &
  \multicolumn{3}{l}{\textit{pride}} &
  \multicolumn{2}{c}{0.42} &
  \multicolumn{2}{c}{0.56} &
  0.48 &
  602 \\ \hline
\textit{anxiety} &
  \multicolumn{2}{c}{0.55} &
  \multicolumn{2}{c}{0.65} &
  \multicolumn{2}{c}{0.59} &
  960 &
  \multicolumn{3}{l}{\textit{arrogance}} &
  \multicolumn{2}{c}{0.44} &
  \multicolumn{2}{c}{0.50} &
  0.47 &
  743 \\ \hline
\textit{distrust} &
  \multicolumn{2}{c}{0.61} &
  \multicolumn{2}{c}{0.78} &
  \multicolumn{2}{c}{0.69} &
  1539 &
  \multicolumn{3}{l}{\textit{surprise}} &
  \multicolumn{2}{c}{0.55} &
  \multicolumn{2}{c}{0.62} &
  0.58 &
  922 \\ \hline
\textit{anger} &
  \multicolumn{2}{c}{0.73} &
  \multicolumn{2}{c}{0.86} &
  \multicolumn{2}{c}{0.79} &
  1538 &
  \multicolumn{3}{l}{\textit{realization}} &
  \multicolumn{2}{c}{0.52} &
  \multicolumn{2}{c}{0.58} &
  0.54 &
  1030 \\ \hline
\textit{gessepany} &
  \multicolumn{2}{c}{0.39} &
  \multicolumn{2}{c}{0.21} &
  \multicolumn{2}{c}{0.27} &
  208 &
  \multicolumn{3}{l}{\textit{resolute}} &
  \multicolumn{2}{c}{0.47} &
  \multicolumn{2}{c}{0.43} &
  0.45 &
  416 \\ \hline
\textit{laziness} &
  \multicolumn{2}{c}{0.39} &
  \multicolumn{2}{c}{0.20} &
  \multicolumn{2}{c}{0.26} &
  290 &
  \multicolumn{3}{l}{\textit{NO EMOTION}} &
  \multicolumn{2}{c}{0.54} &
  \multicolumn{2}{c}{0.59} &
  0.56 &
  725 \\ \hline
\textit{sorrow} &
  \multicolumn{2}{c}{0.41} &
  \multicolumn{2}{c}{0.33} &
  \multicolumn{2}{c}{0.36} &
  263 &
  \multicolumn{3}{l}{} &
  \multicolumn{2}{c}{} &
  \multicolumn{2}{c}{} &
   &
   \\ \hline
\textit{preposterous} &
  \multicolumn{2}{c}{0.70} &
  \multicolumn{2}{c}{0.88} &
  \multicolumn{2}{c}{0.78} &
  2055 &
  \multicolumn{3}{l}{} &
  \multicolumn{2}{c}{} &
  \multicolumn{2}{c}{} &
   &
   \\ \hline
\textit{fed up} &
  \multicolumn{2}{c}{0.46} &
  \multicolumn{2}{c}{0.56} &
  \multicolumn{2}{c}{0.51} &
  816 &
  \multicolumn{3}{l}{\textbf{micro avg}} &
  \multicolumn{2}{c}{0.60} &
  \multicolumn{2}{c}{0.72} &
  0.66 &
  39651 \\ \hline
\textit{compassion} &
  \multicolumn{2}{c}{0.52} &
  \multicolumn{2}{c}{0.57} &
  \multicolumn{2}{c}{0.54} &
  685 &
  \multicolumn{3}{l}{\textbf{macro avg}} &
  \multicolumn{2}{c}{0.54} &
  \multicolumn{2}{c}{0.61} &
  0.56 &
  39651 \\ \hline
\textit{pathetic} &
  \multicolumn{2}{c}{0.64} &
  \multicolumn{2}{c}{0.80} &
  \multicolumn{2}{c}{0.71} &
  1519 &
  \multicolumn{3}{l}{\textbf{weighted avg}} &
  \multicolumn{2}{c}{0.60} &
  \multicolumn{2}{c}{0.72} &
  0.65 &
  39651 \\ \hline
\textit{exhaustion} &
  \multicolumn{2}{c}{0.53} &
  \multicolumn{2}{c}{0.46} &
  \multicolumn{2}{c}{0.49} &
  473 &
  \multicolumn{3}{l}{\textbf{samples avg}} &
  \multicolumn{2}{c}{0.61} &
  \multicolumn{2}{c}{0.75} &
  0.65 &
  39651 \\ \hline\hline
\end{tabular}}

\resizebox{\linewidth}{!}{
\begin{tabular}{lc|llc|llllc|llc|lllc}

\multicolumn{17}{c}{\textbf{AUC}} \\ \hline
\textit{dissatisfaction} &
  0.94 &
  \multicolumn{2}{l}{\textit{embarrassment}} &
  \multicolumn{2}{c|}{0.84} &
  \multicolumn{3}{l}{\textit{irritation}} &
  0.92 &
  \multicolumn{2}{l}{\textit{sadness}} &
  \multicolumn{2}{c|}{0.90} &
  \multicolumn{2}{l}{\textit{despair}} &
  0.84 \\ \hline
\textit{shame} &
  0.74 &
  \multicolumn{2}{l}{\textit{boredom}} &
  \multicolumn{2}{c|}{0.88} &
  \multicolumn{3}{l}{\textit{disappointment}} &
  0.88 &
  \multicolumn{2}{l}{\textit{disgust}} &
  \multicolumn{2}{c|}{0.89} &
  \multicolumn{2}{l}{\textit{shock}} &
  0.84 \\ \hline
\textit{reluctant} &
  0.79 &
  \multicolumn{2}{l}{\textit{fear}} &
  \multicolumn{2}{c|}{0.89} &
  \multicolumn{3}{l}{\textit{contempt}} &
  0.93 &
  \multicolumn{2}{l}{\textit{guilt}} &
  \multicolumn{2}{c|}{0.86} &
  \multicolumn{2}{l}{\textit{anxiety}} &
  0.86 \\ \hline
\textit{distrust} &
  0.87 &
  \multicolumn{2}{l}{\textit{anger}} &
  \multicolumn{2}{c|}{0.94} &
  \multicolumn{3}{l}{\textit{gessapany}} &
  0.84 &
  \multicolumn{2}{l}{\textit{laziness}} &
  \multicolumn{2}{c|}{0.82} &
  \multicolumn{2}{l}{\textit{sorrow}} &
  0.85 \\ \hline
\textit{fed up} &
  0.83 &
  \multicolumn{2}{l}{\textit{preposterous}} &
  \multicolumn{2}{c|}{0.89} &
  \multicolumn{3}{l}{\textit{compassion}} &
  0.87 &
  \multicolumn{2}{l}{\textit{pathetic}} &
  \multicolumn{2}{c|}{0.88} &
  \multicolumn{2}{l}{\textit{exhaustion}} &
  0.85 \\ \hline
\textit{admiration} &
  0.93 &
  \multicolumn{2}{l}{\textit{happiness}} &
  \multicolumn{2}{c|}{0.92} &
  \multicolumn{3}{l}{\textit{joy}} &
  0.93 &
  \multicolumn{2}{l}{\textit{gratitude}} &
  \multicolumn{2}{c|}{0.92} &
  \multicolumn{2}{l}{\textit{excitement}} &
  0.93 \\ \hline
\textit{care} &
  0.89 &
  \multicolumn{2}{l}{\textit{expectancy}} &
  \multicolumn{2}{c|}{0.88} &
  \multicolumn{3}{l}{\textit{comfort}} &
  0.88 &
  \multicolumn{2}{l}{\textit{welcome}} &
  \multicolumn{2}{c|}{0.89} &
  \multicolumn{2}{l}{\textit{interest}} &
  0.87 \\ \hline
\textit{relief} &
  0.89 &
  \multicolumn{2}{l}{\textit{respect}} &
  \multicolumn{2}{c|}{0.92} &
  \multicolumn{3}{l}{\textit{attracted}} &
  0.92 &
  \multicolumn{2}{l}{\textit{pride}} &
  \multicolumn{2}{c|}{0.87} &
  \multicolumn{2}{l}{\textit{arrogance}} &
  0.83 \\ \hline
\textit{surprise} &
  0.85 &
  \multicolumn{2}{l}{\textit{realization}} &
  \multicolumn{2}{c|}{0.83} &
  \multicolumn{3}{l}{\textit{resolute}} &
  0.86 &
  \multicolumn{2}{l}{\textit{NO EMOTION}} &
  \multicolumn{2}{c|}{0.87} &
  \multicolumn{2}{l}{\textbf{macro avg}} &
  0.88 \\ \hline\hline
\multicolumn{17}{c}{\textbf{MCC: 0.588}} \\ \hline\hline
\end{tabular}
}

\caption{\raggedright Performance metrics}
\end{table*}

\clearpage
\twocolumn
\section{Appendix: Ethical Consideration}
\label{sec:appendix_C}

It is well known that a large dataset inevitably has discrimination against protected groups, and the demand of a fair model is not negligible. Our dataset is not an exception. In this section, we point out such problem and instantiate that a simple method helps to alleviate the discrimination. Here, we focus on gender discrimination as an example.

\subsection{Bias Detection}
The very first question is whether the texts in the source data are biased. We collected 3.2m comments for the source data and sampled 50k for KOTE. To detect discrimination, we use comments not used for the learning. The comments that include words referring to protected groups and their counterparts are collected. Since we focus on gender discrimination, the texts containing one of the gender words, \textit{women}, \textit{men}, \textit{female}, and \textit{male}, are collected. Texts that have both genders are removed. 53k and 38k texts are identified to have female words or male words, respectively. 30k texts are randomly sampled from each gender text set for emotion analysis.

\begin{figure*}[!ht]
\centering
\includegraphics[width=16cm]{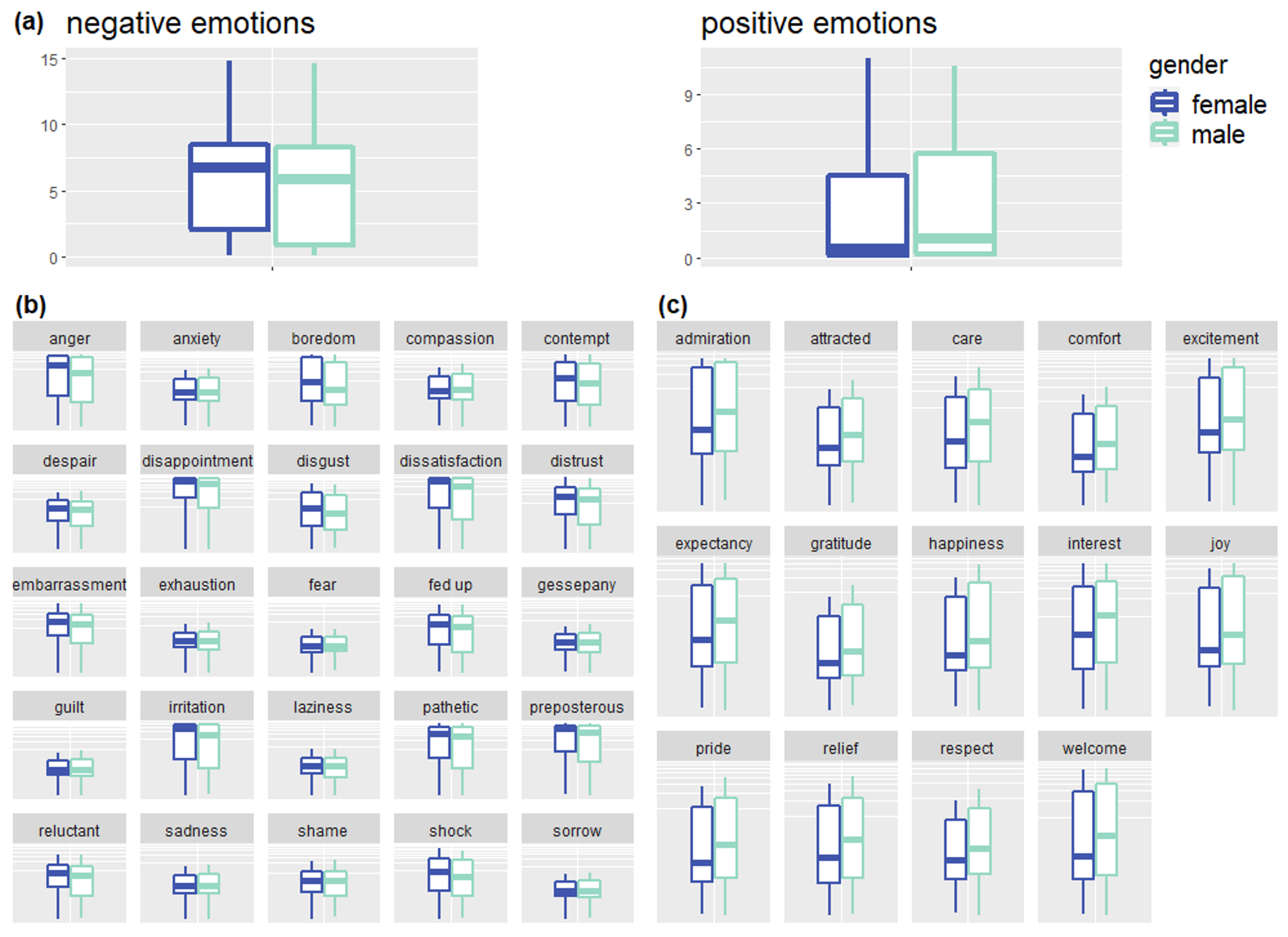}

\caption{\raggedright A comparison of emotions between female and male texts in which the gender tokens are masked. The first plot in (a) compares the sum of negative emotions of each comment in the gender text sets. The second plot in (a) compares the sum of positive emotions of each comment in the gender text sets. In (b) and (c), each box of each plot represents an emotion recognized in the 30k texts. (b) shows how different each negative emotion is by gender, and (c) shows how different each positive emotion is by gender. (b) and (c) are log transformed to illustrate the differences visually.
(plot package; {\fontfamily{cmss}\selectfont ggplot2})
}
\label{fig:3}
\end{figure*}

The texts in both sets are analyzed by the \texttt{KcELECTRA} trained with KOTE, while the gender words are masked with the special token, [MASK]. As in \textbf{Figure~\ref{fig:3}}, the texts containing female words are generally evaluated more negatively, and the texts containing male words are generally evaluated more positively. In conclusion, the source data is biased in the first place, and thus the model could only be biased regardless of the potential discrimination of the raters.

The second question is whether and how much the trained model is biased. To answer this question, we borrow the basic idea of explainable machine learning via token switching. From the source data, we input 320k texts (10\% of the total source data) into the model and select 500 nonoverlapping texts that have the highest probabilities for each label (22k in total). Then, two randomly selected tokens (except [PAD], [CLS], and [SEP]) of each text are replaced with the female words (i.e., \textit{women} and \textit{female}) or the male words (i.e., \textit{men} and \textit{male}). As a result, 22k random-to-female switched texts and 22k random-to-male switched texts are produced. The model would evaluate the two text sets equally if it is fair.

\begin{figure*}[!ht]
\centering
\includegraphics[width=16cm]{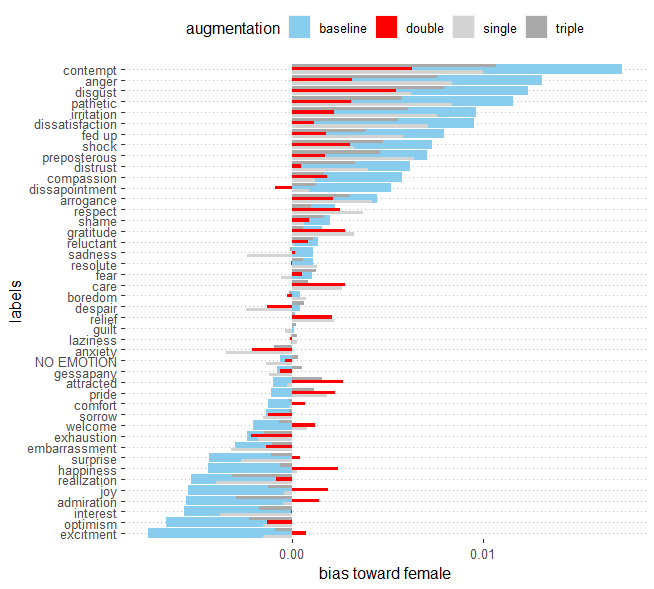}

\caption{\raggedright The bars indicate the mean difference of each label’s probabilities between the texts in which two random tokens are replaced with the female words and the texts in which two random tokens are replaced with the male words. The texts with female words are evaluated more negative. The bias is most serious in the baseline model (the light blue bars). On the other hand, models trained with additional gender-swapped texts are relatively less biased, and the decrease of the bias is largest when the gender-swapped texts as well as the original texts containing gender words are augmented twice (the red bars).
(plot packages; {\fontfamily{cmss}\selectfont ggplot2} and  {\fontfamily{cmss}\selectfont ggpubr}.)
}
\label{fig:4}
\end{figure*}

The results are presented in \textbf{Figure~\ref{fig:4}}. The bars show the mean difference of each label’s predicted probabilities between the two text sets. The light blue bars indicate the baseline model without a manipulation for fairness. The positive direction indicates the bias toward female. The baseline model evaluates the texts more negative on average when some tokens are replaced with the female words. In contrast, the same texts with the male words are evaluated more positive on average. In particular, the texts with the female words are evaluated discriminatorily for negative-intense emotions (e.g., \textit{contempt, anger, disgust, pathetic}, and \textit{irritation}).

\subsection{Unbiasing}

One of the simplest but powerful methods to mitigate discrimination in a language dataset is data augmentation with token switching (\citealp{zhao-etal-2018-gender, park-etal-2018-reducing}). We swap the gender tokens to generate additional texts, and then add the generated texts on the train set.

940 texts in our train set are identified to have at least one gender word. The gender tokens in the texts are replaced with their antonym (\textit{female} to \textit{male}, \textit{women} to \textit{men}, and vice versa) and these gender-swapped texts are added on the original train set to create 40,940 instances in total. Also, we trained a double and triple augmented model, in which the original texts and the gender-swapped texts are augmented one and two more times respectively, in order to accentuate the texts containing the gender tokens.

\textbf{Figure~\ref{fig:4}} shows the results. The augmented models are less biased than the baseline model, and the double augmented model is the least biased. Furthermore, the augmented models cause no critical change in the performance metrics. In the double augmented model, the average F1-score increases by 0.002, the average AUC decreases by 0.0002, and the MCC hardly changes.

Of course, there exist a variety of more thorough methods that help to mitigate biases (For survey and review, see \citealp{sun-etal-2019-mitigating, caton2020fairness, mehrabi2021survey}). However, we would like to emphasize that bias can be alleviated with little attention, and the model performance may not be impaired much. Hence, it is recommended to use a fairer model. Especially, when the dataset is used for a machine designed for direct interaction with humans or other sensitive situations, a strong recommendation is to proceed with caution and go through the process of mitigating discrimination.

\end{document}